\documentclass[10pt,journal,letterpaper]{IEEEtran}
\usepackage{cite}

\ifCLASSINFOpdf
  \usepackage[pdftex]{graphicx}
\else
\fi
\usepackage{amsmath}
\usepackage{amssymb}
\usepackage{array}
\usepackage{booktabs}
\usepackage[table]{xcolor}
\usepackage{hhline}
\usepackage{multirow}
\usepackage{xpatch}

\AtBeginEnvironment{figure}{\setlength{\abovecaptionskip}{3pt}}
\AtBeginEnvironment{figure*}{\setlength{\abovecaptionskip}{3pt}}
\AtEndEnvironment{figure}{\par\vspace{-4pt}}
\AtEndEnvironment{figure*}{\par\vspace{-4pt}}

\newcolumntype{C}[1]{>{\centering\arraybackslash}p{#1}}
\newcommand{\rotlabel}{Rot.}
\newcommand{\translabel}{Trans.}
\usepackage[hidelinks]{hyperref}
\begin{document}
%
\title{Can a Display Be Used for Rapid LiDAR-Camera Extrinsic Calibration?}
%
%
%

\author{Zuhao~Zou,~\IEEEmembership{Graduate Student Member,~IEEE,}
Xiangcheng~Hu,
Jianhao~Jiao,
and Jin~Wu,~\IEEEmembership{Senior Member,~IEEE}%
\thanks{This work was supported by National Key Research and Development Program under the grant of 2025YFB4711400, National Natural Science Foundation of China under the grant of 62603050; Zuhao Zou and Xiangcheng Hu contributed equally to this work. (\emph{Corresponding author: Jin Wu})}
\thanks{Z. Zou is with the Department of Mechanical Engineering, The University of Hong Kong, Pokfulam, Hong Kong (e-mail: zuhao.zou@connect.hku.hk); X. Hu is with the Department of Electronic and Computer Engineering, The Hong Kong University of Science and Technology, Hong Kong (e-mail: xhubd@connect.ust.hk); J. Jiao is with Lion Rock AI Lab (e-mail: jiaojh1994@gmail.com); J. Wu is with the School of Artificial Intelligence, University of Science and Technology Beijing, Beijing, China (e-mail: jin\_wu\_uestc@hotmail.com).}%
}
%
%

\markboth{IEEE Transactions xxx, Submitted Manuscript}%
{Zou \MakeLowercase{\textit{et al.}}: Display-Based LiDAR--Camera Extrinsic Calibration}
%



\maketitle

\begin{abstract}
ScreenCalib enables light detection and ranging (LiDAR)--camera extrinsic calibration using an ordinary programmable display, providing known visual markers and controlled geometry while reducing dedicated target preparation. The camera observes displayed AprilTags, whereas the LiDAR measures the physical screen surface and boundary, requiring indirect cross-modal association. Display geometry introduces association ambiguities, while indoor clutter complicates extraction. ScreenCalib addresses these challenges through tag-guided display extraction, geometric hypothesis generation and refinement, and surrounding scene edges for ambiguity resolution. The single-pose mode requires only one static placement. The joint multi-pose mode combines independent static observations through cross-pose consensus and jointly refines the shared extrinsics and tag scale, reducing sensitivity to tag-size measurement errors without requiring continuous motion or inter-pose trajectory estimation. Experiments in simulation and on three different sensor platforms cover mechanical and nonrepetitive LiDARs in clean and cluttered environments. Benchmark results show that ScreenCalib outperforms competitive calibration-board-based methods in translation accuracy. Joint multi-pose refinement further reduces mean rotation and translation errors across the three platforms and improves rendered-map alignment. Evaluation across readily available displays, including cell phones, tablets, laptops, computer monitors, and televisions, further examines ScreenCalib's versatility as a practical alternative to dedicated calibration targets. The source code is available at \url{https://github.com/ZuhaoZou2026/ScreenCalib}.
\end{abstract}

\begin{IEEEkeywords}
Camera calibration, light detection and ranging (LiDAR), multisensor systems, robot sensing systems.
\end{IEEEkeywords}

%
\IEEEpeerreviewmaketitle

\section{Introduction}
\label{sec:introduction}
\begin{figure*}[t]
    \centering
    \includegraphics[width=\textwidth]{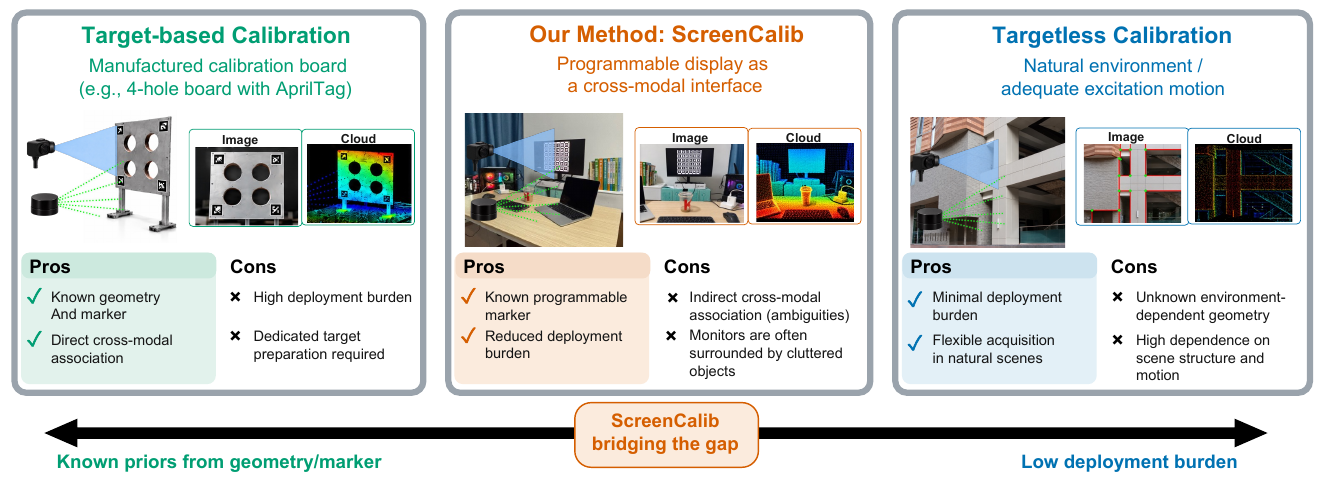}
    \caption{Comparison of target-based, ScreenCalib, and targetless LiDAR--camera calibration. Target-based methods provide known and repeatable geometry and direct cross-modal association, but require dedicated target preparation. Targetless methods minimize deployment burden by exploiting natural scene structure, but the available calibration geometry depends on the environment. ScreenCalib uses an ordinary programmable display to retain known calibration geometry while reducing dedicated target preparation. Its main challenges are the indirect association between camera and LiDAR observations and the cluttered environments in which displays are commonly located.}
    \label{fig:calib_tradeoff}
\end{figure*}

Accurate extrinsic calibration between light detection and ranging (LiDAR) sensors and cameras is fundamental to multimodal perception~\cite{vora2020pointpainting}, multimodal place recognition~\cite{zou_ibtc}, and three-dimensional reconstruction~\cite{jiang2025ligs}. In practical robotic systems, however, calibration is constrained not only by estimation accuracy, but also by how easily reliable calibration observations can be obtained during deployment. Existing methods typically make a trade-off between these two requirements. As illustrated in Fig.~\ref{fig:calib_tradeoff}, target-based methods deliberately introduce known and repeatable geometric structures, but require dedicated target preparation and deployment. Targetless methods substantially reduce this burden, but the available calibration constraints depend on the surrounding environment and sensor motion. When the scene does not provide sufficiently informative geometry or texture, or when the motion excitation is inadequate, the calibration problem can become weakly constrained, leading to unstable estimates, degraded accuracy, or even calibration failure. This raises a straightforward practical question: \emph{can we achieve \textbf{reliable and repeatable} LiDAR--camera calibration \textbf{without carrying and setting up} a dedicated calibration target?}

Target-based LiDAR--camera calibration uses specially designed structures, such as planar boards, circular holes, or checkerboards, to establish explicit cross-modal correspondences~\cite{zhou2018automatic,beltran2022automatic,jiao2023lcecalib}. The known and repeatable target geometry (as priors) generally provides well-constrained and reliable calibration. However, this comes with a relatively high deployment burden, as the target must be fabricated or printed, accurately measured, transported, and positioned for each calibration.

Targetless methods occupy the other end of this trade-off. They avoid dedicated calibration objects by exploiting natural scene appearance, geometric structures, sensor motion, or learned cross-modal correspondences~\cite{yuan2021pixel,chen2025ikalibr}. This reduces deployment burden, but the resulting calibration strongly depends on scene structure and motion excitation; insufficient or repetitive observations can lead to unstable estimates or calibration failure. Some methods also rely on accurate initialization, odometry, or trajectory optimization, further increasing sensitivity to acquisition conditions.

ScreenCalib is designed to reduce this dependency without returning to a specially manufactured calibration target. Instead of carrying and setting up a dedicated board, we use an ordinary display as a \emph{programmable cross-modal calibration interface}. The displayed pattern provides known and repeatable visual geometry to the camera, while the physical display surface and its finite boundary provide geometric structure to the LiDAR. The displayed content can be electronically generated and adjusted according to sensing distance, field of view, and image resolution, without printing, attaching, or machining a dedicated target. As illustrated in Fig.~\ref{fig:calib_tradeoff}, this places ScreenCalib between conventional target-based and targetless calibration: it preserves a user-provided and repeatable calibration structure while substantially reducing the deployment burden associated with a dedicated target.

Importantly, however, a display is not simply an electronic replacement for a conventional calibration board. In conventional target-based calibration, both sensing modalities are typically associated through the same physical target features. In ScreenCalib, the two modalities observe different manifestations of the display. The camera directly observes the displayed AprilTags, whereas the LiDAR generally cannot observe these visual fiducials and instead measures the physical screen surface and boundary. The cross-modal relationship must therefore be established indirectly through their shared display geometry. 
This is the first practical difficulty.

The second one is that, the planar geometry of the display inherently introduces association ambiguities: multiple permutations of the four image and LiDAR corners may produce apparently plausible initial transformations. These characteristics make display-based calibration substantially different from directly applying a conventional calibration-board pipeline.

The third practical difficulty arises from where displays are normally used. Unlike a dedicated calibration target that can be placed in a clean acquisition area, displays are commonly located in offices, laboratories, and other cluttered indoor environments. Nearby objects can generate competing image edges and planar LiDAR clusters, while the display boundary may be sparsely or noisily sampled, especially by nonrepetitive solid-state LiDARs.

Furthermore, ScreenCalib requires a measured AprilTag side length as input, and this measurement may be inaccurate. In single-pose data, LiDAR measurement noise can inflate points near the display boundary, making it difficult to jointly estimate the tag size and extrinsics accurately. This motivates joint multi-pose calibration.

To address these challenges, we propose \textbf{ScreenCalib}, a display-based LiDAR--camera calibration framework with single-pose (SP) and multi-pose (MP) modes. The main contributions are as follows:
\begin{itemize}

    \item \textbf{Robust Display Extraction in Two Different Modalities.}
    We use an ordinary programmable display as a calibration target. On the camera side, AprilTag detections define a reliable polygon around the displayed pattern. Because the pattern and outer display boundaries are parallel in the display plane, the projected pattern edges guide the search for the outer display boundary in the image. 
    The AprilTag-derived depth prior range-filters LiDAR points
    for display-cloud extraction. The detected tags also provide a pixel-to-meter ratio, which, together with the image-polygon size, gives a metric size cue for LiDAR display extraction.
    Experiments in cluttered environments demonstrate the robustness of display extraction.

    \item \textbf{Single-pose calibration with indirect cross-modal association.}
    We develop a complete calibration pipeline requiring only one static display placement. Multiple camera--LiDAR display associations are explicitly retained, initialized from the display geometry, refined by finite-support registration, and validated using surrounding cross-modal scene structures. This enables rapid calibration even when the display is surrounded by cluttered indoor objects.

    \item \textbf{Joint multi-pose calibration.}
    We extend the framework to multiple independently captured static display observations without requiring continuous motion or inter-pose trajectory estimation. Cross-pose consensus selection identifies a consistent initialization from single-pose extrinsic proposals. Joint optimization then combines image-edge, plane, and normal constraints to refine a shared LiDAR--camera transformation and a correction to the user-measured AprilTag size.

    \begin{figure}[!t]
    \centering
    \includegraphics[width=0.9\columnwidth]{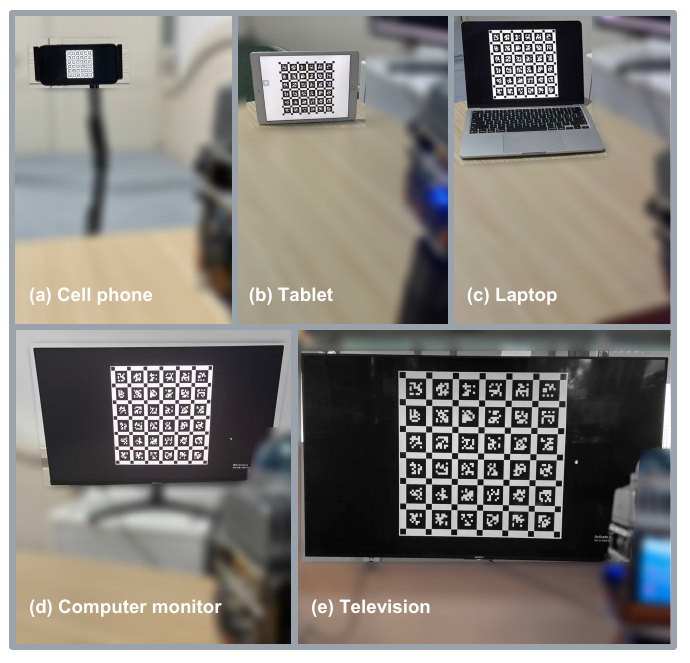}%
    \caption{Examples of display devices used in the ScreenCalib evaluation, each showing the calibration pattern: (a) cell phone, (b) tablet, (c) laptop, (d) computer monitor, and (e) television.}
    \label{fig:test_display}
    \end{figure}

    \item \textbf{Carefully designed experimental evaluation.}
    We examine complementary aspects of the framework: simulation with ground truth; three real LiDAR--camera systems, including mechanical and nonrepetitive LiDARs; clean and cluttered environments; indirect real-world accuracy assessment against Kalibr-estimated inter-camera extrinsics; qualitative assessment through rendered point-cloud alignment; sensitivity to user-measured AprilTag size; and calibration across five display types (Fig.~\ref{fig:test_display}).

\end{itemize}

\begin{figure*}[!t]
  \centering
  \includegraphics[width=1.98\columnwidth]{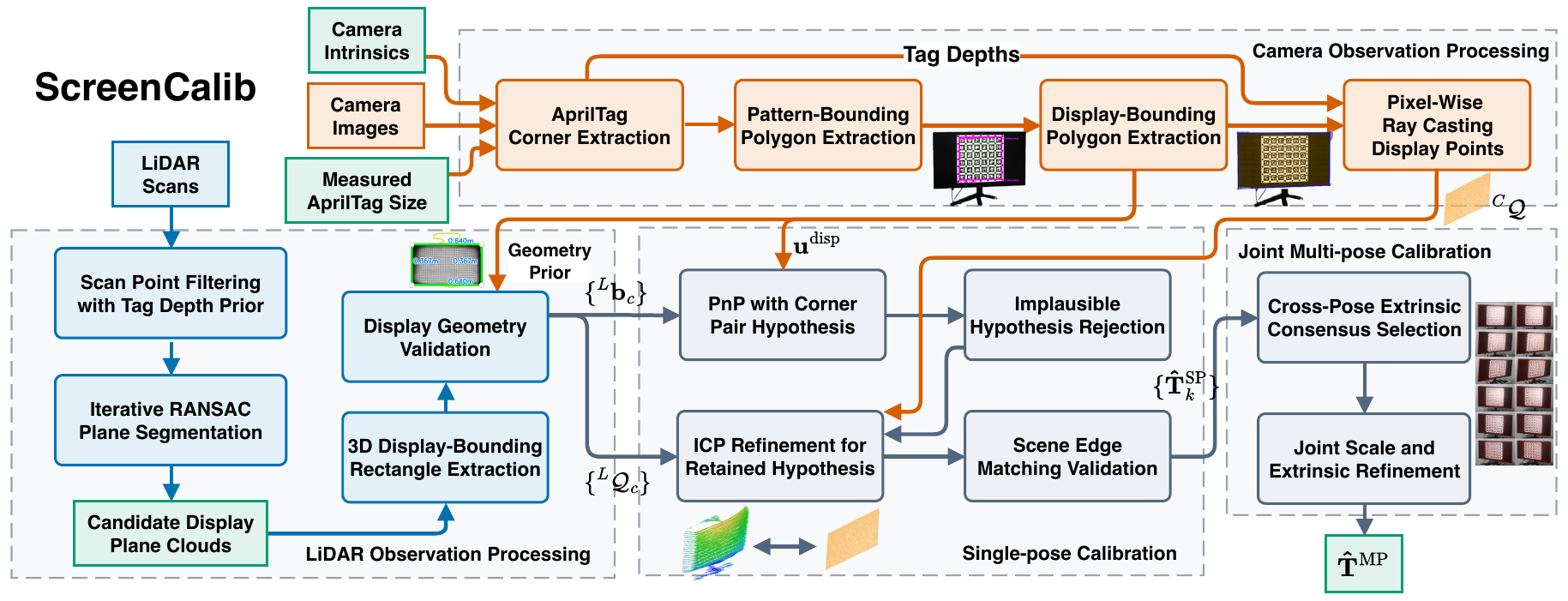}
  \vspace{-0.5em}
  \caption{The ScreenCalib calibration pipeline.}
  \label{fig:pipeline}
\end{figure*}
\vspace{-0.5em}

\section{Related Work}

\subsection{Target-Based Methods}

Target-based LiDAR--camera calibration establishes explicit geometric correspondences using calibration objects with known patterns or structures. Existing approaches differ in their features, preparation requirements, and how targets are isolated from the scene.

Zhou et al.~\cite{zhou2018automatic} combine checkerboard plane and boundary constraints to estimate extrinsics from a single pose, requiring reliable LiDAR edge extraction. Verma et al.~\cite{verma2019automatic} use checkerboard centers and normals, with user-specified board dimensions and spatial filtering bounds.

Beltr\'an et al.~\cite{beltran2022automatic} register a circular-hole target against multilayer LiDAR data using manual spatial bounds.

Dong et al.~\cite{dong2024automatic} address solid-state LiDAR calibration through a multi-target, multi-shot strategy: pseudoimages provide corner correspondences for coarse initialization, followed by refinement using multiple target planes. LCE-Calib~\cite{jiao2023lcecalib} is also a multi-shot method, tracking a checkerboard across frames and accumulating point-to-plane constraints for initialization before refinement with plane and edge constraints. 
These methods thus exploit multiple target observations in different ways: Dong et al. use several patterned targets, whereas LCE-Calib aggregates repeated observations of a tracked checkerboard.

ScreenCalib instead uses an ordinary display, pairing camera-observed AprilTags with the LiDAR-visible screen surface and boundary. Its multi-pose mode combines independently captured static placements through cross-pose consensus and joint refinement of the shared extrinsics and tag scale.

\subsection{Targetless Methods}

Targetless LiDAR--camera calibration avoids dedicated calibration objects by exploiting natural scene appearance, geometric features, or sensor motion. These approaches simplify target preparation, but their applicability depends on the availability of the measurements needed to constrain calibration.

Pandey et al.~\cite{pandey2012automatic} maximize mutual information between LiDAR reflectivity and image intensity, requiring sufficient statistical dependence between the two measurements. Geometric approaches align natural features: Yuan et al.~\cite{yuan2021pixel} use edges formed by intersecting surfaces, while Zhang et al.~\cite{zhang2021line} use line features for online calibration. These methods require reliably associated and sufficiently distributed scene features.

CMRNext~\cite{cattaneo2025cmrnext} predicts image-to-LiDAR correspondences and uncertainties, followed by robust perspective-$n$-point (PnP) estimation, and supports unseen sensor configurations without retraining. MDPCalib~\cite{petek2024automatic} combines odometry-based initialization with learned correspondences and requires reliable odometry and adequate motion.

iKalibr~\cite{chen2025ikalibr} performs targetless spatiotemporal calibration of heterogeneous sensors through dynamic initialization and continuous-time optimization. It requires informative motion and sufficient trajectory observations.

Han et al.~\cite{han2026unicalib} propose UniCalib, which converts camera images and sparse LiDAR points into unified dense depth maps and estimates probabilistic flow to represent cross-modal correspondence uncertainty. A reliability map and perceptually weighted sparse-flow loss suppress unreliable regions, improving robustness to occlusions and dynamic objects.

Compared with targetless methods, ScreenCalib uses an ordinary programmable display to provide known, repeatable geometric constraints, reducing dependence on natural scene structure, texture, and sensor-motion excitation. Thus, ScreenCalib combines the low deployment burden of targetless methods with reliable geometric and marker constraints. It requires neither LiDAR reflectivity nor trajectory estimation. Surrounding-scene geometry helps distinguish competing single-pose hypotheses when the display outline is ambiguous.




\section{Problem Formulation}

The pipeline of the proposed method is shown in Fig.~\ref{fig:pipeline}. Furthermore, this section formulates the problem of estimating the extrinsic transformation of a rigidly mounted LiDAR--camera system from observations of an AprilTag-patterned display. It also introduces the key variables and geometric objects used throughout the method, whose notation is summarized in Tab.~\ref{tab:notations_sec_iv}.

\subsection{Coordinate Frames and Calibration Variables}

Let $L$, $C$, and $M$ denote the LiDAR, camera, and display coordinate frames, respectively. The display frame is attached to the displayed pattern, with its $z$-axis normal to the display plane. A point expressed in frame $A$ is denoted by ${}^{A}\mathbf{p}$, and ${}^{A}\mathbf{T}_B$ maps coordinates from frame $B$ to frame $A$.

The unknown extrinsic is $\mathbf{T}={}^C\mathbf{T}_L=(\mathbf{R},\mathbf{t})\in\mathrm{SE}(3)$, where $\mathbf{R}={}^C\mathbf{R}_L$ and $\mathbf{t}={}^C\mathbf{t}_L$, satisfying
\begin{equation}
{}^{C}\mathbf{p}
= \mathbf{R}{}^{L}\mathbf{p}+\mathbf{t}.
\end{equation}

Camera intrinsics and distortion parameters are calibrated beforehand. All images are undistorted upon receipt, and $\pi_C$ projects points into these undistorted images. The user measures the physical side length $s_0$ of an AprilTag directly on the display to provide the nominal metric scale.

\begin{table}[!t]
\centering
\caption{Key Notations}
\label{tab:notations_sec_iv}
\footnotesize
\renewcommand{\arraystretch}{1.1}
\begin{tabular}{@{}l@{\hspace{4pt}}p{0.82\columnwidth}@{}}
\toprule
Notation & Explanation \\
\midrule
$L$, $C$, $M$ & LiDAR, camera, and display coordinate frames. \\
${}^{A}\mathbf{p}$ & A point expressed in coordinate frame $A$. \\
${}^{A}\mathbf{T}_B$ & Transformation mapping coordinates from frame $B$ to frame $A$. \\
$\mathbf{T}$ & LiDAR-to-camera extrinsic, for simplicity, $\mathbf{T}={}^C\mathbf{T}_L=(\mathbf{R},\mathbf{t})\in\mathrm{SE}(3)$. \\
$\mathbf{R}$, $\mathbf{t}$ & Rotation and translation components of the LiDAR-to-camera extrinsic. \\
$\pi_C$ & Calibrated projection into the undistorted image. \\
$s_0$ & Nominal AprilTag side length used to set metric scale. \\
${}^{C}\mathcal{Q}$ & Virtual display cloud reconstructed in the camera frame. \\
${}^{L}\mathcal{Q}_c$ & The $c$-th candidate display cloud extracted from the LiDAR observations, expressed in the LiDAR frame.\\
$\mathbf{u}^{\mathrm{disp}}_j$ & The $j$-th 2D corner of the display polygon in the image. \\
${}^{L}\mathbf{b}_{c,j}$ & The $j$-th 3D corner of the $c$-th LiDAR display candidate. \\
$c$ & Index of a detected plane candidate fitted from the LiDAR cloud. \\
$h$ & For a given plane candidate, index of a 3D--2D corner-matching hypothesis; each retained PnP branch is treated as a separate hypothesis. \\
$k$ & Index of a single-pose result used in joint multi-pose calibration. \\
\bottomrule
\end{tabular}
\end{table}
\vspace{-0.5em}

\subsection{Calibration Overview}

The inputs are camera images and LiDAR scans, collected under static placement. Camera intrinsics and distortion are calibrated beforehand.

The display is a shared planar target. AprilTag correspondences recover its plane, and boundary extraction yields the four image corners $\{\mathbf{u}^{\mathrm{disp}}_j\}_{j=1}^{4}$. Intersecting rays through this polygon with the tag-derived plane reconstructs the camera-frame display cloud ${}^{C}\mathcal{Q}$. LiDAR plane fitting and clustering produce candidate display clouds $\{{}^{L}\mathcal{Q}_c\}$, whose 3D corners $\{{}^{L}\mathbf{b}_{c,j}\}_{j=1}^{4}$ are paired with the image corners using the permutations in Sec.~\ref{sec:geometric_hypothesis_generation}.

These measurements support extrinsic estimation. Sec.~\ref{single_pose_mode} handles one pose; Sec.~\ref{sec:joint_multi_pose} combines poses and refines the nominal tag size $s_0$.

\section{Display Observation Extraction}
\label{sec:display_observation_extraction}

Fig.~\ref{fig:pipeline} illustrates how camera observations guide LiDAR processing. AprilTag detections recover the display plane and provide a camera-frame depth prior and pixel-to-meter scale. We use the depth prior to range-filter LiDAR points; the detected 2D display polygon, together with the scale, provides a prior on the display's physical width and height. These complementary cues help identify plausible LiDAR display surfaces and produce more reliable calibration measurements.

\subsection{AprilTag-Based Display Plane Estimation}

Corners of the displayed AprilTags~\cite{olson2011tags} are detected using Kalibr~\cite{oth2013rolling,rehder2016extending}. Their identities associate image corners $\mathbf{u}_j^{\mathrm{tag}}$ with nominal pattern coordinates ${}^{M}\mathbf{q}_j$. These correspondences determine the display pose $\mathbf{T}_M={}^C\mathbf{T}_M=(\mathbf{R}_M,\mathbf{t}_M)$ through calibrated PnP, where $\mathbf{R}_M={}^C\mathbf{R}_M$ and $\mathbf{t}_M={}^C\mathbf{t}_M$. The projection model and image points use a consistent distortion convention.

We represent the display plane by $({}^{C}\mathbf{n})^\top{}^{C}\mathbf{p}={}^{C}d$, where ${}^{C}\mathbf{n}$ is a unit normal and ${}^{C}d$ is the signed plane offset. Since the pattern lies on the $z=0$ plane of the display frame $M$, its plane parameters are
\begin{equation}
{}^{C}\mathbf{n}=\mathbf{R}_M\mathbf{e}_3,
\qquad {}^{C}d=({}^{C}\mathbf{n})^\top\mathbf{t}_M,
\end{equation}
where $\mathbf{e}_3=[0,0,1]^\top$. For nominal tag centers ${}^{M}\mathbf{c}_i$, the finite, positive camera-axis depths $z_i=\mathbf{e}_3^\top(\mathbf{R}_M{}^{M}\mathbf{c}_i+\mathbf{t}_M)$ are averaged to obtain $\bar z=N_{\mathrm{tag}}^{-1}\sum_{i=1}^{N_{\mathrm{tag}}}z_i$, where $N_{\mathrm{tag}}$ counts valid centers. This depth prior guides LiDAR filtering in Sec.~\ref{sec:depth_guided_lidar_display_extraction}.

Together with the image boundary extracted next, this plane defines the virtual display cloud ${}^{C}\mathcal{Q}$: camera rays through the detected display polygon are intersected with the plane, as detailed in Sec.~\ref{sec:display_geometry_reconstruction}.

\subsection{Image-Based Display Boundary Extraction}

In undistorted image coordinates, the farthest tag corner in each quadrant about their centroid forms an initial quadrilateral. Invalid quadrilaterals are rejected. Each edge is searched outward by shifting its midpoint along the edge normal. Candidate segments are oriented toward the vanishing point of the corresponding opposite edges when reliable; otherwise, the initial edge is translated along its normal.

Candidates are scored by counting nearby pixels with strong Sobel gradients approximately normal to the segment. A minimum displacement and comparison with the unshifted score suppress the pattern boundary. After non-maximum suppression, the first retained peak restricts the search interval, within which the farthest candidate with sufficient support relative to the best score is selected.

Nearby Canny edges~\cite{canny1986computational} refine the selected lines. Their intersections yield four cyclically ordered display corners $\{\mathbf{u}_j^{\mathrm{disp}}\}_{j=1}^{4}$ enclosing the image region $\Omega$. Polygons failing convexity, image-boundary, or geometric consistency checks are rejected. Perspective can break parallelism, so opposite-edge directions provide only a weak check.

\subsection{Depth-Guided LiDAR Display Extraction}
\label{sec:depth_guided_lidar_display_extraction}

The accumulated LiDAR points are restricted to
\begin{equation}
\alpha_1\bar z<\|{}^{L}\mathbf{p}\|_2\leq\alpha_2\bar z,
\qquad 0<\alpha_1<\alpha_2,
\end{equation}
where $\alpha_1$ and $\alpha_2$ define a coarse range interval for compact sensor configurations. Camera-axis depth and LiDAR range are generally different quantities.

After optional voxel downsampling, iterative random sample consensus (RANSAC)~\cite{fischler1981ransac} extracts candidate planes and removes their inliers from the residual cloud. Euclidean clustering separates disconnected regions within each plane. Components with sufficient support are retained and their planes are refitted, producing candidate display clouds $\{{}^{L}\mathcal{Q}_c\}$ indexed by $c$.
\label{sec:display_geometry_reconstruction}

\subsection{Display Geometry Reconstruction and Validation}

For a pixel $\mathbf{u}\in\Omega$, its calibrated bearing vector ${}^{C}\mathbf{r}(\mathbf{u})$ intersects the display plane at
\begin{equation}
{}^{C}\mathbf{p}(\mathbf{u})=\lambda(\mathbf{u}){}^{C}\mathbf{r}(\mathbf{u}),
\qquad
\lambda(\mathbf{u})=\frac{{}^{C}d}{({}^{C}\mathbf{n})^\top{}^{C}\mathbf{r}(\mathbf{u})}.
\end{equation}
Only finite intersections with $\lambda(\mathbf{u})>0$ and denominators sufficiently far from zero are retained. These intersections form the camera-frame virtual display cloud, ${}^{C}\mathcal{Q}=\{{}^{C}\mathbf{q}_i\}$; applying them to the image corners recovers the display boundary.

For each LiDAR-frame candidate display cloud ${}^{L}\mathcal{Q}_c$, principal component analysis (PCA) defines a local in-plane coordinate system. The farthest projected point in each quadrant about the centroid is selected as a candidate corner. These corners are ordered cyclically and mapped onto the fitted plane, yielding $\{{}^{L}\mathbf{b}_{c,j}\}_{j=1}^{4}$, where ${}^{L}\mathbf{b}_{c,j}$ is the $j$-th corner of the $c$-th candidate display expressed in the LiDAR frame. Candidate rectangles must have sufficient extent, approximately parallel opposite edges, and side lengths within the prescribed display-size bounds.

The plane $({}^{C}\mathbf{n},{}^{C}d)$, image corners $\{\mathbf{u}_j^{\mathrm{disp}}\}$, reconstructed cloud ${}^{C}\mathcal{Q}$, and LiDAR clouds $\{{}^{L}\mathcal{Q}_c\}$ with their corners $\{{}^{L}\mathbf{b}_{c,j}\}$ are the inputs to the subsequent extrinsic estimation in Secs.~\ref{single_pose_mode} and~\ref{sec:joint_multi_pose}. The alternating 3D edge lengths of these corners provide candidate display width and height, which can be checked for consistency with the 2D corner spans using the pixel-to-meter scale estimated from the AprilTag, helping reject incorrect LiDAR clouds.

\section{Single-Pose Extrinsic Calibration}
\label{single_pose_mode}
The rectangular display outline does not uniquely determine the corner association, and several LiDAR plane candidates may satisfy the geometric checks. We therefore generate alternative hypotheses, reject implausible solutions, and refine the remaining transformations before final selection. Fig.~\ref{fig:hypothesis_rejection} contrasts a correct hypothesis with three incorrect cases addressed by the rejection and validation steps below. The acquisition index is omitted in this section.

\subsection{Geometric Hypothesis Generation}
\label{sec:geometric_hypothesis_generation}

Both the 3D LiDAR candidate corners $\{{}^{L}\mathbf{b}_{c,j}\}_{j=1}^{4}$ and the 2D image polygon vertices $\{\mathbf{u}^{\mathrm{disp}}_j\}_{j=1}^{4}$ are cyclically ordered. However, they leave the cross-modal starting corner and traversal direction ambiguous.

We enumerate four cyclic shifts under both forward and reversed traversal, producing eight adjacency-preserving corner permutations. Each plane candidate and corner permutation defines a 3D--2D association. We use $h$ to index the resulting hypotheses across all plane candidates; any retained PnP branches are indexed separately. Hypotheses are initialized and checked as described below.

\subsection{Extrinsic Initialization and Hypothesis Rejection}

For each hypothesis $h$ of the $c$-th display candidate, using its corner pair correspondence, the extrinsic is initialized by solving the PnP problem:
\begin{equation}
\begin{aligned}
\mathbf{T}^{\mathrm{PnP}}_{h}
&=\arg\min_{\mathbf{T}}\\
&\quad\sum_{j=1}^{4}
\left\|
\pi_C\left(\mathbf{R} \!\!\!\quad{}^{L}\mathbf{b}_{c,j}
+\mathbf{t}\right)
-\mathbf{u}^{\mathrm{disp}}_{\sigma(j)}
\right\|_2^2.
\end{aligned}
\end{equation}

\begin{figure}[!t]
\centering
\includegraphics[width=0.95\columnwidth]{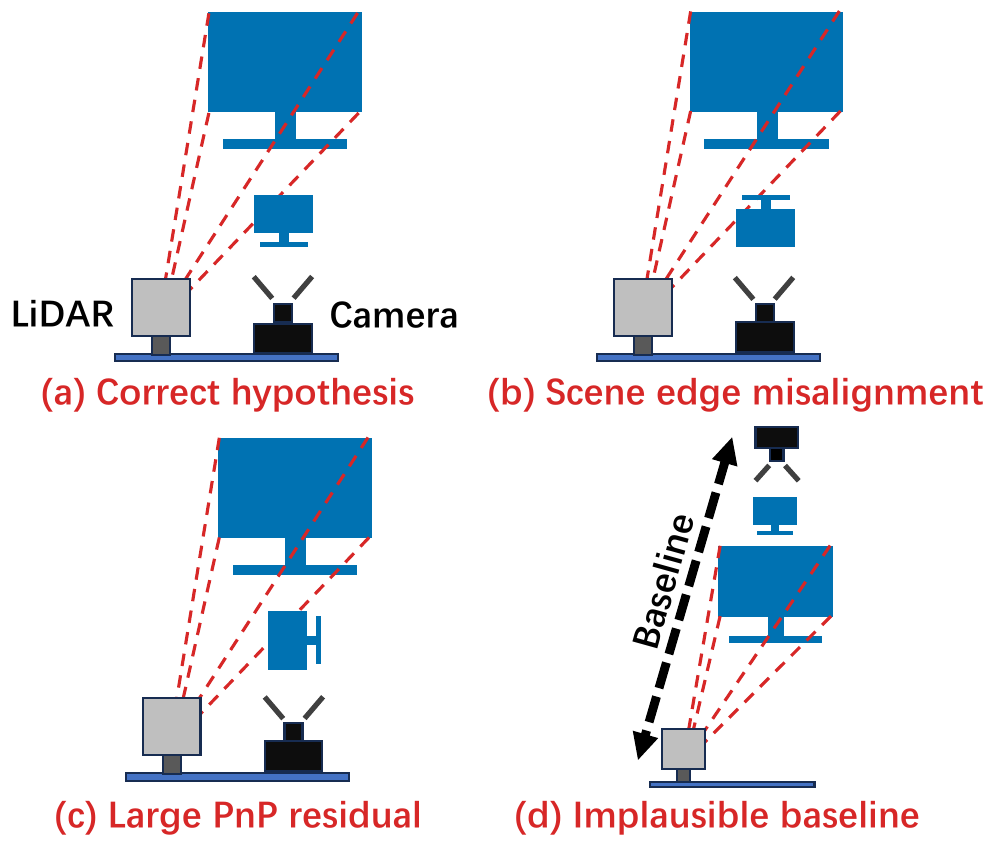}
\caption{Hypothesis cases: (a) correct hypothesis, (b) scene-edge misalignment, (c) large initial PnP residual, and (d) implausible LiDAR--camera baseline.}
\label{fig:hypothesis_rejection}
\end{figure}
\vspace{-0.5em}

Reject any hypothesis with a corner behind the camera.

After the positive-depth check, hypothesis rejection uses two criteria: the initial PnP reprojection residual and the implied LiDAR--camera separation. The dimensionless reprojection error $\epsilon_{\mathrm{rep},h}$ is the root-mean-square pixel error over the four correspondences, divided by the mean image-polygon edge length $\bar\ell$. Hypotheses with $\epsilon_{\mathrm{rep},h}>\tau_{\mathrm{rep}}$ are discarded, as illustrated in Fig.~\ref{fig:hypothesis_rejection}(c), where the proposed corner association yields a large initial reprojection residual.

We also require $\|\mathbf{t}^{\mathrm{PnP}}_{h}\|_2\leq B_{\max}$, where $B_{\max}$ is a coarse sensor-separation bound supplied from the physical configuration. This rejects implausibly large baselines, such as the case in Fig.~\ref{fig:hypothesis_rejection}(d), without specifying the translation direction or an initial rotation. It can eliminate some front--back association errors, but does not guarantee rejection of every incorrect hypothesis.

For notational simplicity, the hypothesis index $h$ is omitted during geometric refinement, where each retained hypothesis is processed independently. It is reinstated in scene-edge validation to compare the retained hypotheses.

\subsection{Geometric Refinement}

Each retained transformation is refined using iterative closest point (ICP) registration of the virtual display cloud reconstructed in the camera frame ${}^{C}\mathcal{Q}$ with its associated LiDAR cloud, denoted by ${}^{L}\mathcal{Q}$ for the current hypothesis. 

For the ICP refinement below, $m=0,1,\ldots$ denotes the iteration index, and $\mathbf{T}^{0}$ is initialized by the PnP estimate above,
i.e., $\mathbf{T}^{0} = \mathbf{T}^{\mathrm{PnP}}$.

The ICP algorithm alternates between nearest-neighbor correspondence estimation and rigid-transformation updates. At iteration $m$, each transformed source point ${}^{C}\mathbf{q}_i\in{}^{C}\mathcal{Q}$ is matched to its nearest target point ${}^{L}\mathbf{q}_i$ in ${}^{L}\mathcal{Q}$. Matches exceeding a prescribed distance threshold are excluded, leaving valid correspondence indices $\mathcal{A}_m$. With these correspondences fixed, the rigid increment is estimated as
\begin{equation}
\begin{aligned}
\Delta\mathbf{T}^{m}
&=\arg\min_{\Delta\mathbf{T}^{m}}\\
&\;\sum_{i\in\mathcal{A}_m}
\left\|
\Delta\mathbf{R}^{m}
\left(\mathbf{R}^{m}\right)^{\top}
\left({}^{C}\mathbf{q}_i-\mathbf{t}^{m}\right)
+\Delta\mathbf{t}^{m}-{}^{L}\mathbf{q}_i
\right\|_2^2.
\end{aligned}
\end{equation}
Here, $\Delta\mathbf{T}^{m}=(\Delta\mathbf{R}^{m},\Delta\mathbf{t}^{m})$ acts on points in the LiDAR frame. Let $\mathbf{T}^{m}=(\mathbf{R}^{m},\mathbf{t}^{m})$ denote the current LiDAR-to-camera estimate; its inverse maps camera points into the LiDAR frame. The left-multiplicative update is $(\mathbf{T}^{m+1})^{-1}=\Delta\mathbf{T}^{m}(\mathbf{T}^{m})^{-1}$, and iteration continues until the increment is sufficiently small or the iteration limit is reached. After ICP convergence, the refined LiDAR-to-camera estimate is $\mathbf{T}^{\mathrm{ICP}}$.

The finite display boundaries provide in-plane alignment information, whose strength depends on boundary coverage and point sampling. ICP convergence alone therefore does not establish a correct association or strong constraints in every degree of freedom. Candidates must converge with sufficient correspondence support and acceptable mean squared alignment error. Sensor-separation and positive-depth checks are repeated after refinement.

\subsection{Scene Edge Matching Validation}
\label{sec:scene_edge_validation}

Using a common rectangular display as an example, its LiDAR cloud may align similarly well with
the reconstructed camera-frame display cloud under both the true extrinsic transformation and an alternative hypothesis differing by a $180^\circ$ rotation about the display normal. A square display introduces additional ambiguities at $90^\circ$ and $270^\circ$.
Display geometry alone may therefore be insufficient to distinguish these hypotheses. Fig.~\ref{fig:hypothesis_rejection}(a)--(b) illustrates this distinction: the correct hypothesis is consistent with the display and surrounding scene, whereas an incorrect hypothesis can retain nice display alignment while misaligning scene edges. We therefore evaluate the alignment between surrounding-scene depth edges and image edges to distinguish the remaining hypotheses.

For each geometrically valid hypothesis $h$, the accumulated LiDAR cloud is projected into the image using its refined transformation $\mathbf{T}^{\mathrm{ICP}}_{h}$ and the calibrated projection model $\pi_C$. A depth image $D_h$ is constructed by retaining the nearest positive camera-axis depth at each projected pixel. Depth discontinuities between valid neighboring pixels form a binary depth-edge map $E_h^D$. Boundaries caused by missing depth and edges within the display region $\Omega$, expanded by a seven-pixel margin, are excluded. Independently, Gaussian smoothing followed by Canny detection produces the binary image-edge map $E^I$.

Let $\mathcal{E}_h^D$ denote the eligible depth-edge pixels in $E_h^D$. The surrounding-scene alignment cost is
\begin{equation}
\label{equ:edge_distance_cost}
\begin{aligned}
S_h &= \frac{1}{|\mathcal{E}_h^D|}
       \sum_{\mathbf{u}\in\mathcal{E}_h^D} d(\mathbf{u}),\\
d(\mathbf{u}) &=
\begin{cases}
\|\mathbf{u}-\mathbf{v}_{\mathbf{u}}\|_2,
& \text{if a match is found},\\
\tau_d, & \text{otherwise}.
\end{cases}
\end{aligned}
\end{equation}
Here, $\mathbf{v}_{\mathbf{u}}$ is the nearest image-edge pixel in $E^I$ within $\tau_d$ pixels of $\mathbf{u}$. Thus, $d(\mathbf{u})$ is the capped image-edge distance, independent of hypothesis $h$; $h$ only affects the eligible depth-edge set.
The threshold $\tau_d$ defines both the maximum nearest-point search distance and the penalty for unmatched edge pixels; 
by default, it is set to $20$ pixels. Therefore, lower costs indicate better alignment. 

\begin{figure}[!t]
\centering
\includegraphics[width=0.75\columnwidth]{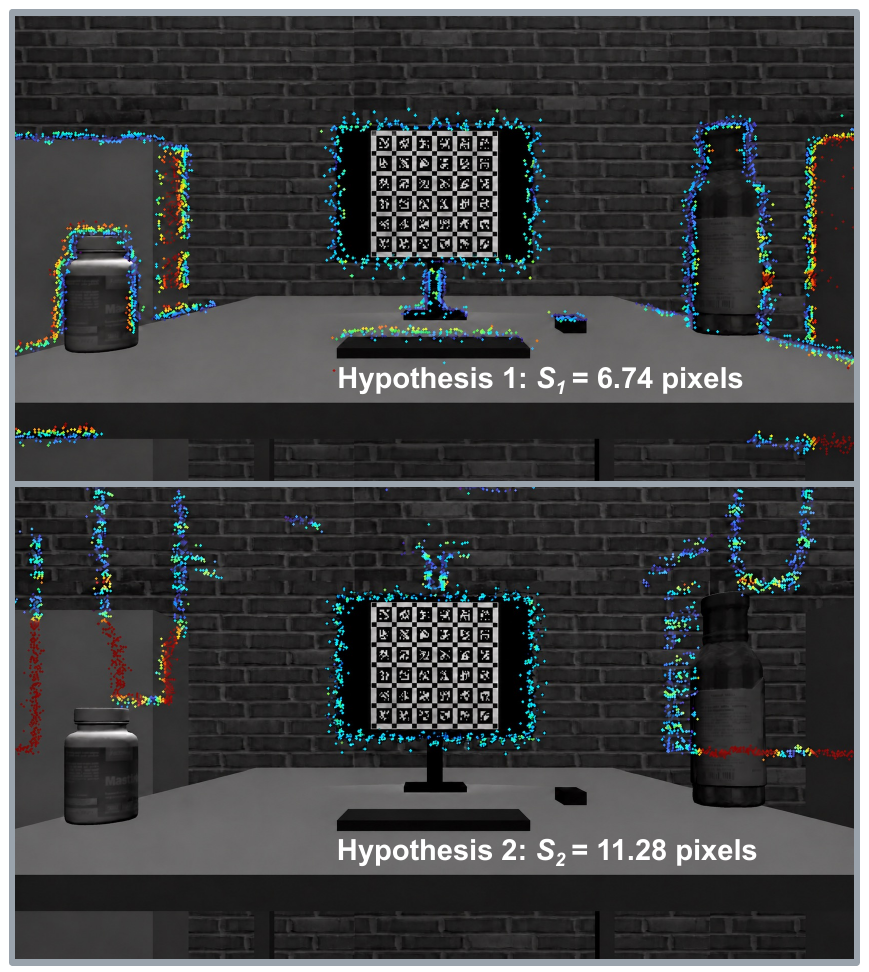}
\caption{Surrounding-scene depth-edge and image-edge alignment for two competing hypotheses. Hypothesis 1 has a lower alignment cost ($S_h=6.74$ pixels) than hypothesis 2 ($S_h=11.28$ pixels), so it is selected.}
\label{fig:hypothesis_edge_alignment}
\end{figure}

The final single-pose estimate is selected as
\begin{equation}
\label{equ:icp_result}
h^*=\arg\min_{h\in\mathcal{H}} S_h,
\qquad
\widehat{\mathbf{T}}^{\mathrm{SP}}
=\mathbf{T}^{\mathrm{ICP}}_{h^*},
\end{equation}
where $\mathcal{H}$ contains the hypotheses that pass initialization, geometric refinement, and validation checks.
Fig.~\ref{fig:hypothesis_edge_alignment} illustrates how the surrounding-scene alignment cost distinguishes two geometrically plausible hypotheses.

\section{Joint Multi-Pose Extrinsic Calibration}
\label{sec:joint_multi_pose}

Single-pose (SP) calibration estimates the extrinsic from one static acquisition. Joint multi-pose (MP) calibration combines static acquisitions of the same rigid sensor pair to refine shared extrinsics and AprilTag scale. Sensors remain stationary within each acquisition but may be repositioned between poses. Secs.~\ref{sec:display_observation_extraction} and~\ref{single_pose_mode} provide a validated display cloud and initial proposal for every pose.

\subsection{Cross-Pose Extrinsic Consensus Selection}
\label{sec:cross_pose_consensus}

Individual proposals may be inaccurate because of imperfect display extraction, geometric ambiguity, or tag-size error. We select a consensus extrinsic initialization by comparing the proposals across poses, then jointly refine it using observations from all acquired poses.

For pose $k$, let $\widehat{\mathbf{T}}_{k} = (\widehat{\mathbf{R}}_{k}, \widehat{\mathbf{t}}_{k})$ be its single-pose estimate (Equ.~\ref{equ:icp_result}); this estimate serves as the pose's proposal.

The $j$-th pose supports the $i$-th pose if their rotational and translational differences are both within the corresponding consistency thresholds ($i \neq j$). Accordingly, the support set of the $i$-th pose is defined as
\begin{equation} 
\begin{gathered} 
\mathcal{S}_i = \left\{ j \,\middle|\, j\neq i \ \land\ \theta_{ij}\leq\tau_R \ \land\ \delta_{ij}\leq\tau_t \right\};\\
\theta_{ij} = \left\| \operatorname{Log} \left( \widehat{\mathbf{R}}_{i}^{\top} \widehat{\mathbf{R}}_{j} \right)^{\vee} \right\|_2
,
\delta_{ij} = \left\| \widehat{\mathbf{t}}_{i} - \widehat{\mathbf{t}}_{j} \right\|_2. \\
\end{gathered} 
\label{eq:cross_pose_support} 
\end{equation}
where $\tau_R$ and $\tau_t$ are the rotation and translation consistency thresholds, respectively. By default, $\tau_R$ is $1^\circ$ and $\tau_t$ is $2$~cm.

If no proposal has a supporting pose, calibration terminates and reports an error asking the user to check whether the target display is fully observed by both the camera and LiDAR. 

We select the proposal with the most supporters. For the same support number case, select the smallest supporting score $s_i$ one:
\begin{equation}
s_i=\frac{1}{|\mathcal{S}_i|}\sum_{j\in\mathcal{S}_i}
\left[\left(\frac{\theta_{ij}}{\tau_R}\right)^2+
\left(\frac{\delta_{ij}}{\tau_t}\right)^2\right].
\end{equation}

A pose does not support itself. The consensus set $\mathcal{C}=\mathcal{S}_{\mathrm{seed}}\cup\{\mathrm{seed}\}$ is used to select the initialization, and its members need not be mutually consistent. The selected consensus extrinsic initializes joint refinement, which also uses observations from poses outside $\mathcal{C}$. 
A pose may fail to support the consensus because the wrong hypothesis survives, even when the display itself was correctly detected. The consensus extrinsic provides an initialization that allows these outlier poses to contribute their local edge and point-plane constraints during joint optimization.

\subsection{Tag Scale Correction}

Let $s_0$ be the user-measured AprilTag side length. We model its correction by $\beta=\exp(\eta)>0$, initialized at $\beta=1$, so the corrected side length is $s=\beta s_0$. The displayed AprilTag retains the same physical side length across all acquisitions used in one joint calibration.

Because $s_0$ sets the metric scale of the virtual display cloud, its measurement error scales the reconstructed points and can bias the extrinsic estimate.

The scale correction is applied to each camera-derived display point ${}^{C}\mathbf{q}_{i} \in {}^{C}\mathcal{Q}$ as
\begin{equation}
{}^{C}\mathbf{q}_{i}(\beta)
=
\beta\,{}^{C}\mathbf{q}_{i},
\end{equation}
This adjusts each reconstructed point's depth along its pixel's viewing ray while preserving the ray direction. The display-plane normal remains unchanged, and its offset scales by $\beta$.

\subsection{Joint Extrinsic and Scale Refinement}
\label{sec:joint_extrinsic_scale_refinement}

The proposed joint MP method estimates $\mathbf{T}=(\mathbf{R},\mathbf{t})$ and shared scale $\beta$ using image-edge, point-to-plane, and plane-normal terms. Using the residuals defined below, the four named objective components are
\begin{equation}
\begin{aligned}
F_{\mathrm{Alignment}}&=\frac{1}{|\mathcal V|}\sum_{k\in\mathcal V}\frac{1}{|\mathcal E_k|}\sum_{i\in\mathcal E_k}\rho\!\left((r^{\mathrm{Alignment}}_{ki})^2\right),\\
F_{\mathrm{ICP}}&=\frac{1}{|\mathcal V|}\sum_{k\in\mathcal V}\frac{1}{|\mathcal M_k|}\sum_{j\in\mathcal M_k}\rho\!\left((r^{\mathrm{ICP}}_{kj})^2\right),\\
F_{\mathrm{Normal}}&=\frac{1}{|\mathcal V|}\sum_{k\in\mathcal V}\rho\!\left(\|\mathbf{r}^{\mathrm{Normal}}_k\|_2^2\right),\\
F_{\mathrm{TagSizePrior}}&=\left(\frac{\beta-1}{\sigma_\beta}\right)^2,\\
F_{\mathrm{Joint}}&=F_{\mathrm{Alignment}}+F_{\mathrm{ICP}}\\
&\!\!\!\!\!\quad+F_{\mathrm{Normal}}+F_{\mathrm{TagSizePrior}}.
\end{aligned}
\end{equation}
The joint calibration result is obtained by solving
\begin{equation}
\left(\widehat{\mathbf{T}}^{\mathrm{MP}},\widehat{\beta}\right)
=\underset{\mathbf{T},\,\beta}{\operatorname{arg\,min}}
\quad F_{\mathrm{Joint}}(\mathbf{T},\beta).
\label{equ:joint_minimization}
\end{equation}
For pose $k$, $\mathcal E_k$ is the set of dynamically associated depth--image edge pairs that pass the pixel-distance gate, and $\mathcal M_k$ is the set of dynamically associated point--plane correspondences that pass the distance gate. Their cardinalities normalize the corresponding sums, and $\rho$ is the Huber robust loss~\cite{huber1964robust}.

All acquired poses are considered for joint refinement, regardless of membership in the consensus set. Correspondence gates select measurements consistent with the current extrinsic estimate, limiting the influence of inconsistent observations. 
Here, $\mathcal V$ denotes poses with sufficient valid gated correspondences to evaluate the data terms.

In the objective above, the indexed residuals with $k$ are expanded as follows with $k$ omitted:
\begin{align}
r^{\mathrm{Alignment}}_{i}&=\frac{d\!\left(\pi_K(\mathbf{R}\mathbf{e}_{i}+\mathbf{t})\right)}{\sigma_{\mathrm{e}}},\\
r^{\mathrm{ICP}}_{j}&=\frac{({}^{C}\mathbf{n})^{\mathsf T}(\mathbf{R}\mathbf{p}_{j}+\mathbf{t})-\beta\,{}^{C}d}{\sigma_{\mathrm{ICP}}},\\
\mathbf{r}^{\mathrm{Normal}}&=\frac{\mathbf{R}\,{}^{L}\mathbf{n}-{}^{C}\mathbf{n}}{\sigma_n},
\end{align}
Here, $d$ denotes the capped image-edge distance $d_k$ defined in Equ.~\ref{equ:edge_distance_cost} from the Canny edge map for pose $k$, and $\pi_K$ denotes the undistorted camera projection. The samples $\mathbf{p}_j$ and $\mathbf{e}_i$ are expressed in the LiDAR frame, and the plane parameters are pose-specific. The vectors ${}^{C}\mathbf{n}$ and ${}^{L}\mathbf{n}$ are the unit display-plane normals in the camera and LiDAR frames, respectively, and ${}^{C}d$ is the signed plane offset in the camera frame. The extensively-validated and fixed positive scales $\sigma_{\mathrm{e}}$, $\sigma_n$, $\sigma_{\mathrm{ICP}}$, and $\sigma_\beta$ normalize the four residual terms.

The four terms provide complementary constraints:
\begin{itemize}
    \item {\textbf{Cross-Modal Edge Alignment:}} Starting from the consensus extrinsic, we project the LiDAR points into a projective depth image, detect edges across the full image area in both this depth image and the RGB image, and penalize the local distance between the two edge sets, following the construction approach in Sec.~\ref{sec:scene_edge_validation}.
    \item {\textbf{Point-to-Plane ICP:}} This term keeps transformed LiDAR display points on the virtual plane.
    \item {\textbf{Plane Normal Difference:}} This term refines extrinsic orientation by aligning the transformed LiDAR normal with the camera-plane normal.
    \item {\textbf{Tag Size Prior:}} This term allows the tag-size scale $\beta$ to be optimized, adjusting the depth of each virtual display point along its pixel's viewing ray, while regularizing $\beta$ toward $1$.
\end{itemize}

\textit{\textbf{Remark 1:}}
Edge-inflated LiDAR points can bias single-pose scale and extrinsic estimates. Complementary static poses can reduce this bias during joint extrinsic and tag-scale refinement.

\section{Experiments}
We evaluate ScreenCalib through both simulated and real-world experiments, which are designed to benchmark ScreenCalib's single-pose (SP) and multi-pose (MP) modes in terms of accuracy, robustness, and practical versatility. Because simulated sequences provide exact ground-truth extrinsics and controllable scene conditions, they offer the clearest quantitative evidence of calibration accuracy. Although real-world sequences lack exact LiDAR--camera ground truth, they expose the methods to real sensor noise and enable accuracy comparisons across different LiDAR--camera configurations. Finally, tests across various display types and controlled AprilTag-size errors examine whether ScreenCalib remains effective with different displays and imperfectly measured AprilTag sizes.

We compare ScreenCalib with Zhou et al.~\cite{zhou2018automatic}, Beltr\'an et al.~\cite{beltran2022automatic}, two highly cited works, and LCE-Calib~\cite{jiao2023lcecalib}, a recent multi-pose calibration method. We use the publicly available GitHub implementations of Beltr\'an et al. and LCE-Calib, and Zhou et al.'s MATLAB implementation. To ensure a fair comparison, we use the same simulated and real-world data collection environments, place the calibration boards and displays in similar configurations, and use boards with the default designs and dimensions reported in their publications.

Zhou et al., Beltr\'an et al., and the proposed SP variant use a single static pose (single-shot calibration), whereas LCE-Calib and the proposed MP variant use multiple static poses (multi-shot calibration).

\subsection{Calibration Accuracy Benchmark on Simulated Data}

The simulated environments provide ground-truth LiDAR--camera extrinsics, allowing both rotational and translational calibration errors to be evaluated. Tab.~\ref{tab:simulated_calibration_error} compares the proposed single-pose (SP) and joint multi-pose (MP) variants with the baseline methods for Ouster--camera and Velodyne--camera configurations. In both configurations, noise with a standard deviation of 1.0~cm is added to the LiDAR points.

Tab.~\ref{tab:simulated_calibration_error} reports SP rotation/translation errors of 0.36$^\circ$/0.66~cm (Ouster) and 0.28$^\circ$/0.51~cm (Velodyne). SP gives the lowest single-shot translation errors in both cases and the lowest Ouster rotation error; Beltr\'an et al. slightly outperform it on Velodyne (0.25$^\circ$ vs. 0.28$^\circ$). MP errors are 0.34$^\circ$/0.40~cm and 0.20$^\circ$/0.27~cm, respectively. MP gives the lowest translation errors in both cases and the lowest Ouster rotation error, while LCE-Calib leads on Velodyne rotation (0.14$^\circ$).

\begin{table}[!t]
\caption{Rotation and Translation Calibration Errors on Simulated Sequences}
\label{tab:simulated_calibration_error}
\centering
\scriptsize
\setlength{\tabcolsep}{0.5pt}
\renewcommand{\arraystretch}{1.15}
\begin{tabular*}{\columnwidth}{@{}p{0.55in}|@{\extracolsep{\fill}}C{0.55in}C{0.55in}C{0.55in}|C{0.55in}C{0.55in}@{}}
\hline
\multicolumn{6}{c}{\textbf{Device 1: Ouster (noise standard deviation: 1.0~cm) + Gazebo camera}} \\
\hline
\textbf{Error}
& \shortstack{\textbf{Zhou}\\\textbf{et al.}}
& \shortstack{\textbf{Beltr\'an}\\\textbf{et al.}}
& \shortstack{\textbf{Ours}\\\textbf{(SP)}}
& \shortstack{\textbf{LCE-}\\\textbf{Calib}}
& \shortstack{\textbf{Ours}\\\textbf{(MP)}} \\
\hline
\rotlabel~($^\circ$) & \cellcolor{green!10}1.25 & \cellcolor{green!10}0.52 & \cellcolor{green!10}0.36 & \cellcolor{green!10}0.39 & \cellcolor{green!10}\textbf{0.34} \\
\translabel~(cm) & \cellcolor{blue!10}1.35 & \cellcolor{blue!10}1.71 & \cellcolor{blue!10}0.66 & \cellcolor{blue!10}1.45 & \cellcolor{blue!10}\textbf{0.40} \\
\hline
\end{tabular*}
\vspace{1mm}
\begin{tabular*}{\columnwidth}{@{}p{0.55in}|@{\extracolsep{\fill}}C{0.55in}C{0.55in}C{0.55in}|C{0.55in}C{0.55in}@{}}
\hline
\multicolumn{6}{c}{\textbf{Device 2: Velodyne (noise standard deviation: 1.0~cm) + Gazebo camera}} \\
\hline
\textbf{Error}
& \shortstack{\textbf{Zhou}\\\textbf{et al.}}
& \shortstack{\textbf{Beltr\'an}\\\textbf{et al.}}
& \shortstack{\textbf{Ours}\\\textbf{(SP)}}
& \shortstack{\textbf{LCE-}\\\textbf{Calib}}
& \shortstack{\textbf{Ours}\\\textbf{(MP)}} \\
\hline
\rotlabel~($^\circ$) & \cellcolor{green!10}0.32 & \cellcolor{green!10} 0.25 & \cellcolor{green!10}0.28 & \cellcolor{green!10}\textbf{0.14} & \cellcolor{green!10}0.20 \\
\translabel~(cm) & \cellcolor{blue!10}1.81 & \cellcolor{blue!10} 1.83 & \cellcolor{blue!10}0.51 & \cellcolor{blue!10}0.32 & \cellcolor{blue!10}\textbf{0.27} \\
\hline
\end{tabular*}

\parbox{\columnwidth}{\scriptsize $\cdot$ $\times$ denotes that the method failed to produce a valid calibration result.}
\parbox{\columnwidth}{\scriptsize $\cdot$ Beltr\'an et al. requires manually-tuned calibration-cloud bounds along all three axes.}
\end{table}
\vspace{-0.5em}

\subsection{Calibration Accuracy Benchmark on Real-World Data}
\label{sec:real_world_accuracy}

\begin{figure}[!t]
\centering
\includegraphics[width=\columnwidth]{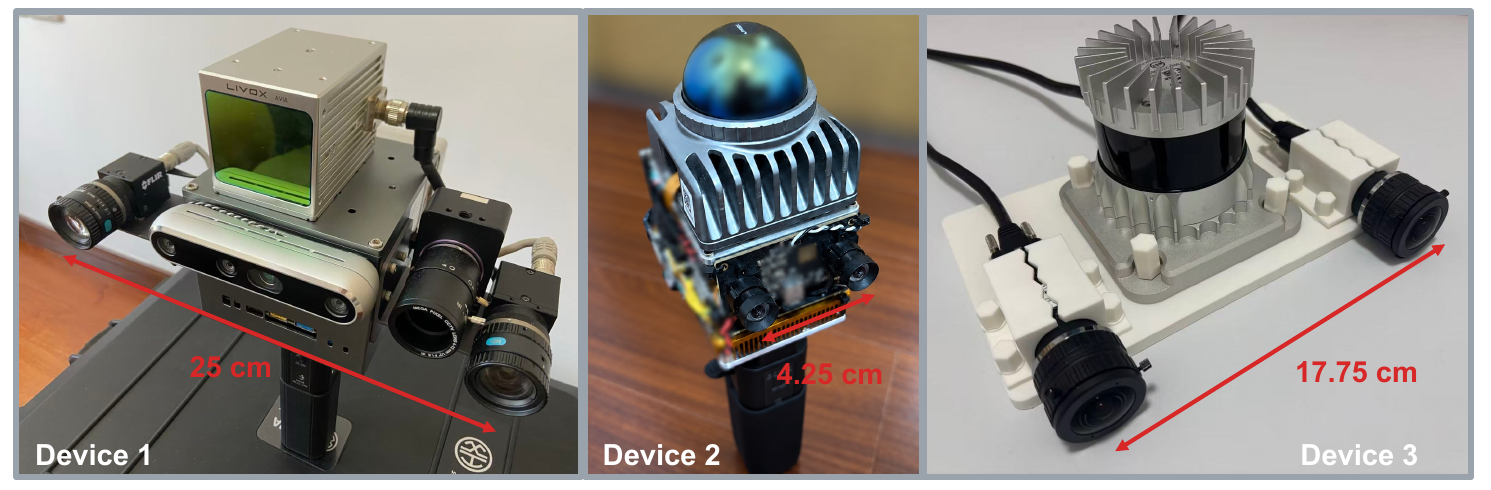}
\caption{The three real-world LiDAR--camera systems evaluated in Tab.~\ref{tab:real_world_translation_error}: Device 1, Livox Avia with two FLIR cameras; Device 2, Livox Mid-360 with two USB cameras; and Device 3, Ouster 64 with two Hikvision cameras. Red arrows indicate the camera baselines and their CAD lengths.}
\label{fig:test_devices}
\end{figure}

Tab.~\ref{tab:real_world_translation_error} compares rotation (degrees) and translation (centimeters) errors for the proposed method and three baselines across real-world sequences. The three LiDAR--camera systems are shown in Fig.~\ref{fig:test_devices}; Device 2 uses USB cameras.

\textit{\textbf{Ground Truth}}. Real-world sequences lack ground-truth LiDAR--camera extrinsics. CAD models provide only nominal poses: manufacturing tolerances, assembly errors, or base bending may shift the actual sensor poses. We therefore do not treat CAD extrinsics as ground truth.

Following prior work~\cite{fu2023extrinsic,bai2020lidar,giacomini2024ca2lib}, we use Kalibr-estimated inter-camera extrinsics~\cite{furgale2013kalibr} as the reference and assess LiDAR--camera accuracy through inter-camera error.

To assess the reliability of the Kalibr results as a reference for indirect evaluation, we compare the estimated camera baselines with the CAD values. As shown in Tab.~\ref{tab:kalibr_cad_comparison}, the Kalibr camera baselines differ from the CAD values by only 0.03, 0.01, and 0.13~cm for Devices 1, 2, and 3, respectively. For each method, we compose its two LiDAR--camera transformations through the shared LiDAR frame and compute the resulting inter-camera rotational and translational errors relative to the Kalibr reference.

\textit{\textbf{Benchmark}}. According to Tab.~\ref{tab:real_world_translation_error}, neither Beltr\'an et al.'s method nor LCE-Calib is applicable to the Livox Avia or Mid-360 LiDAR. In particular, Beltr\'an et al.'s publication and released implementation assume that ``measurements distributed into different layers, as typical in mechanical devices based on rotating mirrors''~\cite{beltran2022automatic}.

\begin{table}[!t]
\caption{Inter-Camera Rotation and Translation Errors on Real-World Sequences From Three LiDAR--Camera Devices}
\label{tab:real_world_translation_error}
\centering
\scriptsize
\setlength{\tabcolsep}{0.5pt}
\renewcommand{\arraystretch}{1.15}
\begin{tabular*}{\columnwidth}{@{}p{0.29in}p{0.46in}|@{\extracolsep{\fill}}C{0.51in}C{0.51in}C{0.51in}|C{0.51in}C{0.51in}@{}}
\hline
\multicolumn{7}{c}{\textbf{Device 1: Livox Avia + 2 FLIR cameras}} \\
\hline
\textbf{Env.} & \textbf{Error}
& \shortstack{\textbf{Zhou}\\\textbf{et al.}}
& \shortstack{\textbf{Beltr\'an}\\\textbf{et al.}}
& \shortstack{\textbf{Ours}\\\textbf{(SP)}}
& \shortstack{\textbf{LCE-}\\\textbf{Calib}}
& \shortstack{\textbf{Ours}\\\textbf{(MP)}} \\
\hline
Clean & \rotlabel~($^\circ$) & \cellcolor{green!10}4.29, 7.84 & \cellcolor{green!10}$\times$ & \cellcolor{green!10}0.81, 0.96 & \cellcolor{green!10}$\times$ & \cellcolor{green!10}0.85, 0.99 \\
Messy & \rotlabel~($^\circ$) & \cellcolor{green!10}3.03, 4.14 & \cellcolor{green!10}$\times$ & \cellcolor{green!10} 2.92, 1.15 & \cellcolor{green!10}$\times$ & \cellcolor{green!10}1.22, 1.76 \\
\hhline{--|---|--}
Clean & \translabel~(cm) & \cellcolor{blue!10}6.35, 29.01 & \cellcolor{blue!10}$\times$ & \cellcolor{blue!10}3.55, 3.95  & \cellcolor{blue!10}$\times$ & \cellcolor{blue!10}1.90, 2.55 \\
Messy & \translabel~(cm) & \cellcolor{blue!10}11.80, 9.91 & \cellcolor{blue!10}$\times$ & \cellcolor{blue!10}11.32, 4.54 & \cellcolor{blue!10}$\times$ & \cellcolor{blue!10}2.12, 3.47 \\
\hhline{--|---|--}
Mean & \rotlabel~($^\circ$) & \cellcolor{green!10}4.83 & \cellcolor{green!10}$\times$ & \cellcolor{green!10}1.46 & \cellcolor{green!10}$\times$ & \cellcolor{green!10}\textbf{1.21} \\
     & \translabel~(cm) & \cellcolor{blue!10}14.27 & \cellcolor{blue!10}$\times$ & \cellcolor{blue!10}5.84 & \cellcolor{blue!10}$\times$ & \cellcolor{blue!10}\textbf{2.51} \\
\hline
\end{tabular*}
\begin{tabular*}{\columnwidth}{@{}p{0.29in}p{0.46in}|@{\extracolsep{\fill}}C{0.51in}C{0.51in}C{0.51in}|C{0.51in}C{0.51in}@{}}
\hline
\multicolumn{7}{c}{\textbf{Device 2: Livox Mid-360 + 2 USB cameras}} \\
\hline
\textbf{Env.} & \textbf{Error}
& \shortstack{\textbf{Zhou}\\\textbf{et al.}}
& \shortstack{\textbf{Beltr\'an}\\\textbf{et al.}}
& \shortstack{\textbf{Ours}\\\textbf{(SP)}}
& \shortstack{\textbf{LCE-}\\\textbf{Calib}}
& \shortstack{\textbf{Ours}\\\textbf{(MP)}} \\
\hline
Clean & \rotlabel~($^\circ$) & \cellcolor{green!10}4.29, 5.50 & \cellcolor{green!10}$\times$ & \cellcolor{green!10}0.46, 0.36 & \cellcolor{green!10}$\times$ & \cellcolor{green!10}0.15, 0.45 \\
Messy & \rotlabel~($^\circ$) & \cellcolor{green!10}62.03, 51.09 & \cellcolor{green!10}$\times$ & \cellcolor{green!10}0.34, 0.37 & \cellcolor{green!10}$\times$ & \cellcolor{green!10}0.43, 0.14 \\
\hhline{--|---|--}
Clean & \translabel~(cm) & \cellcolor{blue!10}6.63, 1.81 & \cellcolor{blue!10}$\times$ & \cellcolor{blue!10}0.25, 0.42  & \cellcolor{blue!10}$\times$ & \cellcolor{blue!10}0.19, 0.91 \\
Messy & \translabel~(cm) & \cellcolor{blue!10}48.91, 21.98 & \cellcolor{blue!10}$\times$ & \cellcolor{blue!10}0.72, 0.56 & \cellcolor{blue!10}$\times$ & \cellcolor{blue!10}0.42, 0.24 \\
\hhline{--|---|--}
Mean & \rotlabel~($^\circ$) & \cellcolor{green!10}30.73 & \cellcolor{green!10}$\times$ & \cellcolor{green!10}0.38 & \cellcolor{green!10}$\times$ & \cellcolor{green!10}\textbf{0.29} \\
     & \translabel~(cm) & \cellcolor{blue!10}19.83 & \cellcolor{blue!10}$\times$ & \cellcolor{blue!10}0.49 & \cellcolor{blue!10}$\times$ & \cellcolor{blue!10}\textbf{0.44} \\
\hline
\end{tabular*}
\vspace{1mm}
\begin{tabular*}{\columnwidth}{@{}p{0.29in}p{0.46in}|@{\extracolsep{\fill}}C{0.51in}C{0.51in}C{0.51in}|C{0.51in}C{0.51in}@{}}
\hline
\multicolumn{7}{c}{\textbf{Device 3: Ouster 64 + 2 Hikvision cameras}} \\
\hline
\textbf{Env.} & \textbf{Error}
& \shortstack{\textbf{Zhou}\\\textbf{et al.}}
& \shortstack{\textbf{Beltr\'an}\\\textbf{et al.}}
& \shortstack{\textbf{Ours}\\\textbf{(SP)}}
& \shortstack{\textbf{LCE-}\\\textbf{Calib}}
& \shortstack{\textbf{Ours}\\\textbf{(MP)}} \\
\hline
Clean & \rotlabel~($^\circ$) & \cellcolor{green!10}3.74, 5.78 & \cellcolor{green!10}2.47, 2.56 & \cellcolor{green!10}3.36, 4.30 & \cellcolor{green!10}3.40, 3.51 & \cellcolor{green!10}3.27, 2.07 \\
Messy & \rotlabel~($^\circ$) & \cellcolor{green!10}36.60, 72.87 & \cellcolor{green!10}3.14, 2.94 & \cellcolor{green!10}1.32, 3.55 & \cellcolor{green!10}14.36, 5.12 & \cellcolor{green!10}1.84, 3.98 \\
\hhline{--|---|--}
Clean & \translabel~(cm) & \cellcolor{blue!10}48.88, 37.64 & \cellcolor{blue!10}6.93, 6.57 & \cellcolor{blue!10}2.44, 3.48 & \cellcolor{blue!10}1.73, 1.22 & \cellcolor{blue!10}2.44, 1.88 \\
Messy & \translabel~(cm) & \cellcolor{blue!10}13.17, 33.31 & \cellcolor{blue!10}5.73, 6.67 & \cellcolor{blue!10}2.58, 2.97 & \cellcolor{blue!10}32.92, 5.75 & \cellcolor{blue!10}3.69, 1.92 \\
\hhline{--|---|--}
Mean  & \rotlabel~($^\circ$) & \cellcolor{green!10}29.75 & \cellcolor{green!10}\textbf{2.78} & \cellcolor{green!10}3.13 & \cellcolor{green!10}6.60 & \cellcolor{green!10}2.79 \\
      & \translabel~(cm) & \cellcolor{blue!10}33.25 & \cellcolor{blue!10}6.48 & \cellcolor{blue!10}2.87 & \cellcolor{blue!10}10.41 & \cellcolor{blue!10}\textbf{2.48} \\
\hline
\end{tabular*}

\parbox{\columnwidth}{\scriptsize $\cdot$ $\times$ denotes that the method failed to produce a valid calibration result.}
\parbox{\columnwidth}{\scriptsize $\cdot$ For each device, we collect four sequences: two in clean environments and two in cluttered (messy) environments. The reported means are the arithmetic averages over these four sequences.}
\parbox{\columnwidth}{\scriptsize $\cdot$ Beltr\'an et al. requires manually-tuned calibration-cloud bounds along all three axes.}
\end{table}

\begin{table}[!t]
\caption{Kalibr Inter-Camera Extrinsics and Comparison with CAD Design (Translation in Centimeters; Rotation in Degrees)}
\label{tab:kalibr_cad_comparison}
\centering
\scriptsize
\setlength{\tabcolsep}{0.5pt}
\renewcommand{\arraystretch}{1.15}
\resizebox{\columnwidth}{!}{%
\begin{tabular}{l|cccc|c|cccc|c}
\hline
& \multicolumn{4}{c|}{Kalibr}
& \multicolumn{1}{c|}{CAD}
& \multicolumn{4}{c|}{Kalibr}
& \multicolumn{1}{c}{CAD} \\
& $t_x$ & $t_y$ & $t_z$ & baseline & baseline & roll & pitch & yaw & angle & angle \\
\cline{2-11}
\hline
Device 1 & 25.00 & 0.14 & 1.34 & \underline{25.03} & \underline{25.00} & 0.49 & -0.45 & 0.28 & \underline{0.72} & \underline{0.00} \\
Device 2 & 4.24 & 0.01 & -0.02 & \underline{4.24} & \underline{4.25} & 0.10 & 0.86 & -0.01 & \underline{0.86} & \underline{0.00} \\
Device 3 & 17.62 & -0.02 & 0.08 & \underline{17.62} & \underline{17.75} & -0.52 & 0.14 & -1.30 & \underline{1.41} & \underline{0.00} \\
\hline
\end{tabular}%
}
\vspace{1mm}

\end{table}

Using Kalibr as the reference, our MP variant achieves the lowest mean translation error among the three baseline methods and our SP variant across all three devices (2.51, 0.44, and 2.48~cm). It also achieves the lowest mean rotation error on Devices 1 and 2 (1.21$^\circ$ and 0.29$^\circ$); on Device 3, its mean rotation error (2.79$^\circ$) is nearly identical to Beltr\'an et al.'s lowest result (2.78$^\circ$).

Considering only the single-shot methods, our SP variant achieves the lowest mean translation error across all three devices (5.84, 0.49, and 2.87~cm) and the lowest mean rotation error on Devices 1 and 2 (1.46$^\circ$ and 0.38$^\circ$). On Device 3, its mean rotation error (3.13$^\circ$) is slightly higher than Beltr\'an et al.'s (2.78$^\circ$), but substantially lower than Zhou et al.'s (29.75$^\circ$).

\textit{\textbf{Calibration Environment}}. Tab.~\ref{tab:real_world_translation_error} also shows that nearby clutter affects the baseline methods to varying degrees. For example, Zhou et al.'s method exhibits large rotation errors on the Mid-360 and Ouster systems, while LCE-Calib shows increased errors on cluttered Ouster sequences. In contrast, the proposed SP and MP variants generally maintain low errors across clean and cluttered trials, indicating greater robustness to objects surrounding the calibration display.

\subsection{Rendered Map Quality Benchmark on Real-World Data}

\begin{figure}[t]
\centering

\includegraphics[width=0.95\columnwidth]{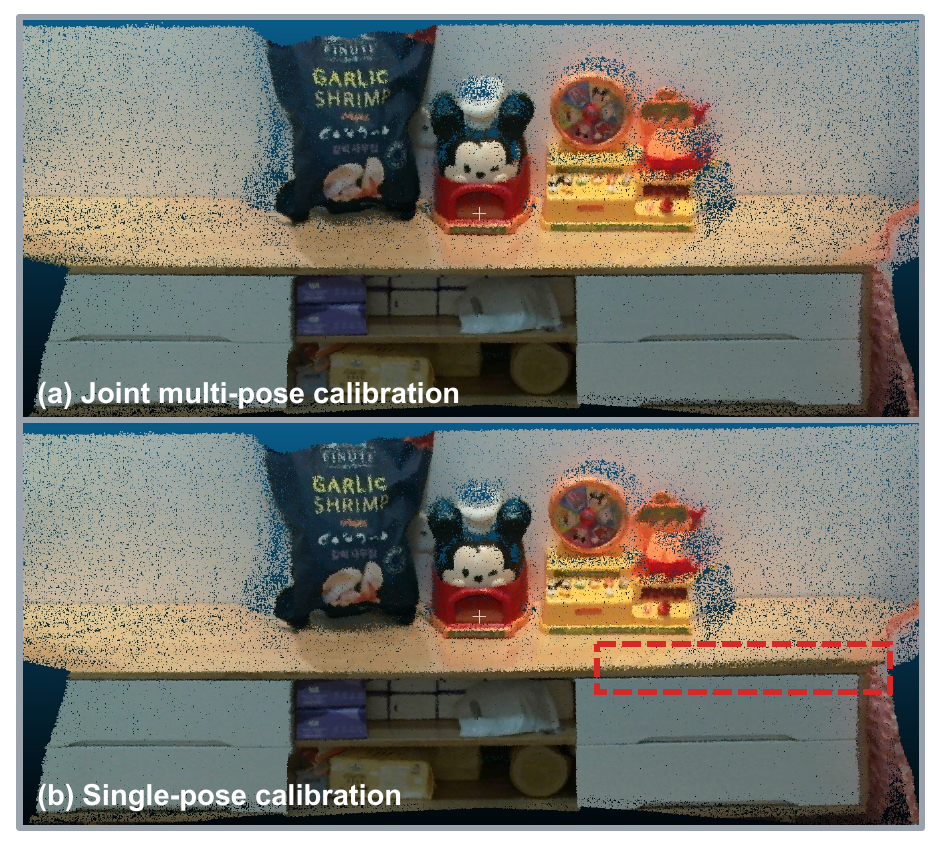}%

\caption{Qualitative comparison of rendered maps generated using the extrinsic calibration parameters estimated by our single-pose (SP) and joint multi-pose (MP) methods: (a) MP and (b) SP. The red box in (b) marks a region with slightly poorer alignment in the SP result. Maps are rendered from real-world data collected by Device 2 in an indoor scene.}
\label{fig:rendered_map_quality}
\end{figure}

This experiment compares the two variants of our method: single-pose (SP) calibration and joint multi-pose (MP) calibration. We render Device 2 data using the extrinsic parameters estimated by each variant; Fig.~\ref{fig:rendered_map_quality}(a) shows the MP result, and Fig.~\ref{fig:rendered_map_quality}(b) shows the SP result. The red box in Fig.~\ref{fig:rendered_map_quality}(b) highlights a region where the SP map is slightly less well aligned, illustrating the qualitative improvement provided by MP. This observation is consistent with the lower mean rotation and translation errors achieved by MP in Tab.~\ref{tab:real_world_translation_error}.

\subsection{Robustness Across Programmable Displays}

We test five displays (Fig.~\ref{fig:test_display}) using the three Device 2 trials in Tab.~\ref{tab:real_world_translation_error}; the errors in Tab.~\ref{tab:display_translation_error} use the Kalibr inter-camera reference (Sec.~\ref{sec:real_world_accuracy}).

Tab.~\ref{tab:display_translation_error} shows that our method in SP mode can provide reasonable calibration results with the tablet, laptop, computer monitor, and television. However, the cell phone does not provide usable results in all three trials, likely because the LiDAR noise at its edges is significant compared with its small size. Among the displays, the television, which has the largest display area, provides the best mean calibration: 0.16$^\circ$ in rotation and 0.30~cm in translation.

Displays support single-capture calibration, but small phones may be less reliable.

\begin{table}[t]
\caption{Single-Pose (SP) Inter-Camera Rotation and Translation Errors Across Displays}
\label{tab:display_translation_error}
\centering
\scriptsize
\setlength{\tabcolsep}{1pt}
\renewcommand{\arraystretch}{1.2}
\setlength{\fboxsep}{1pt}
\newcommand{\rottranscell}[2]{\colorbox{green!10}{#1} / \colorbox{blue!10}{#2}}
\begin{tabular*}{\columnwidth}{@{\extracolsep{\fill}}C{0.90in}C{0.37in}|C{0.50in}C{0.50in}C{0.50in}|C{0.50in}@{}}
\hline
\multicolumn{1}{c}{\multirow{2}{*}{\shortstack{\textbf{Display}}}} & \multicolumn{1}{c|}{\multirow{2}{*}{\shortstack{\textbf{Size}\\\textbf{(cm$\times$cm)}}}} & \multicolumn{3}{c|}{\textbf{Rot. ($^\circ$) / Trans. (cm)}} & \multicolumn{1}{c}{\multirow{2}{*}{\textbf{Mean}}} \\
& & \multicolumn{1}{c}{\textbf{Trial 1}} & \multicolumn{1}{c}{\textbf{Trial 2}} & \multicolumn{1}{c|}{\textbf{Trial 3}} & \\
\hline
Cell phone & 16$\times$7 & \rottranscell{4.81}{1.44} & \rottranscell{$\times$}{$\times$} & \rottranscell{$\times$}{$\times$} & \rottranscell{$\times$}{$\times$} \\
Tablet & 25$\times$18 & \rottranscell{0.12}{0.12} & \rottranscell{2.49}{3.53} & \rottranscell{1.63}{0.61} & \rottranscell{1.41}{1.42} \\
Laptop & 30$\times$20 & \rottranscell{0.23}{0.28} & \rottranscell{0.10}{0.23} & \rottranscell{0.47}{0.44} & \rottranscell{0.27}{0.32} \\
Computer monitor & 70$\times$50 & \rottranscell{0.46}{0.34} & \rottranscell{0.36}{0.42} & \rottranscell{0.37}{0.56} & \rottranscell{0.40}{0.44} \\
Television & 168$\times$96 & \rottranscell{0.13}{0.45} & \rottranscell{0.27}{0.10} & \rottranscell{0.07}{0.35} & \rottranscell{\textbf{0.16}}{\textbf{0.30}} \\
\hline
\end{tabular*}
\parbox{\columnwidth}{\scriptsize $\times$ denotes that the method failed to produce a valid calibration result.}
\end{table}
\vspace{-0.5em}

\subsection{Sensitivity to User-Measured AprilTag Size}

We test tag-size robustness on a simulated sequence with known extrinsics and true AprilTag size. The input size ranges from $-10\%$ to $+10\%$ in $2\%$ steps; for each MP run, we record calibration errors and the optimized $\beta$.

\begin{figure}[h!]
\centering
\includegraphics[width=0.95\columnwidth]{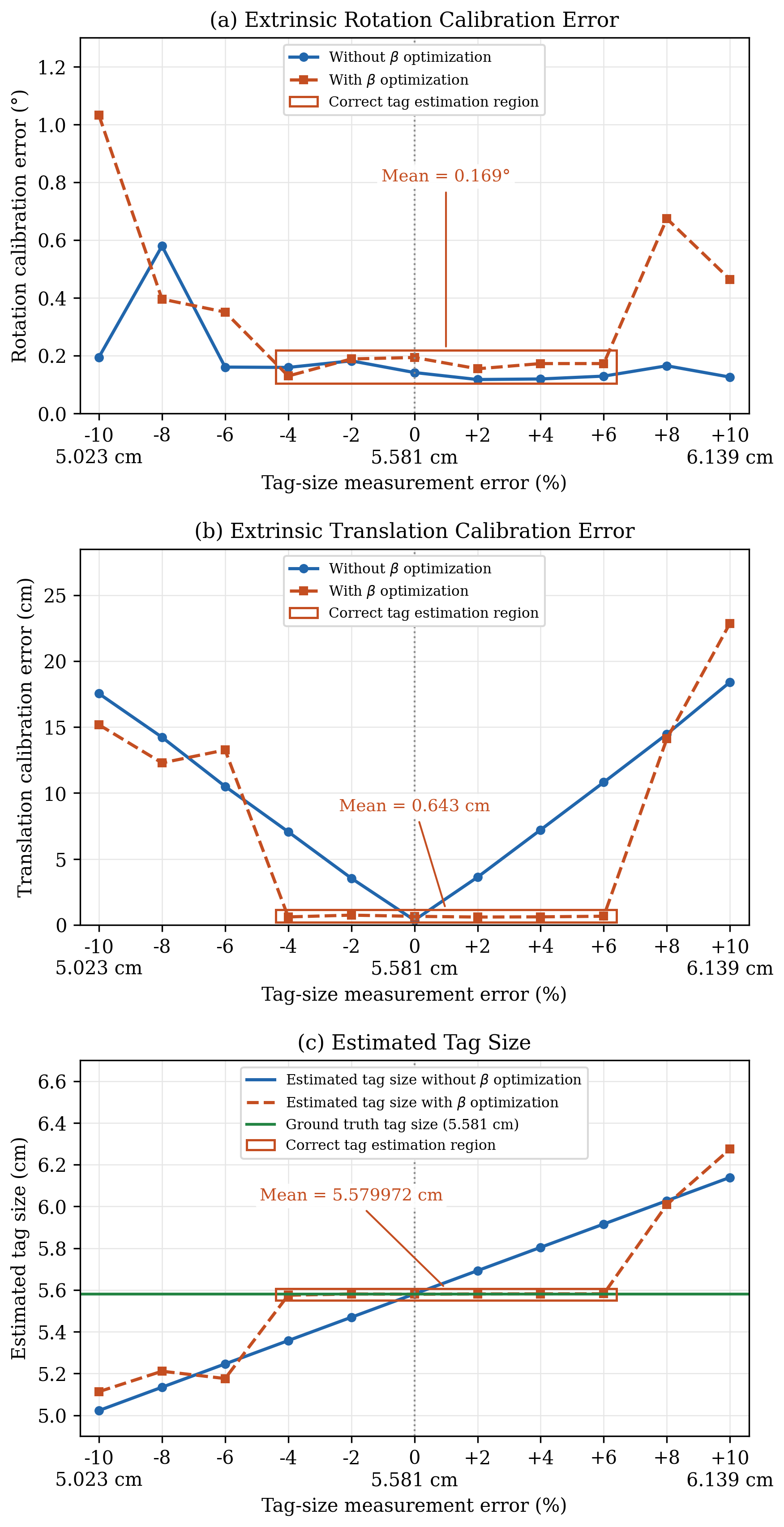}%
\vspace{-0.5em}
\caption{Sensitivity of the multi-pose (MP) method to AprilTag-size measurement errors: (a) extrinsic rotation error, (b) extrinsic translation error, and (c) estimated tag size, shown with and without $\beta$ optimization.}
\label{fig:tag_size_sensitivity}
\end{figure}

Fig.~\ref{fig:tag_size_sensitivity} plots (a) extrinsic rotation error, (b) translation error, and (c) estimated tag size. Plots (a) and (b) show calibration error versus input tag-size error; (c) compares estimated and true tag size. The orange boxes mark the $-4\%$ to $+6\%$ input-error range, where $\beta$ keeps the scale close to ground truth.

Our method performs best for input tag-size errors from $-4\%$ to $+6\%$, with mean rotation and translation errors of $0.169^\circ$ and $0.643$~cm. It remains valid beyond this range, but larger tag-size errors reduce accuracy.

Plot (b) also shows that disabling $\beta$ optimization (blue curve) mainly affects the estimated extrinsic translation. This is because the tag-size scale factor $\beta$ and extrinsic translation along the camera optical axis are coupled.

\vspace{-0.9em}

\section{Conclusion}
This paper presented ScreenCalib, a LiDAR--camera extrinsic calibration framework that uses an ordinary programmable display to retain controlled and repeatable calibration geometry while reducing dedicated target preparation and handling. Its central challenge is that the camera observes the displayed AprilTags, whereas the LiDAR measures the physical screen surface and boundary. ScreenCalib connects these different observations through shared display geometry: tag-guided image-boundary extraction reconstructs the camera-frame display, while tag-derived depth and metric size cues guide LiDAR display extraction in cluttered environments. To address ambiguous corner associations, the single-pose mode retains competing hypotheses, refines them through finite-support registration, and uses surrounding scene edges to distinguish geometrically plausible alternatives. The multi-pose mode combines independently captured static observations through cross-pose consensus and jointly refines the shared extrinsics and tag scale, reducing sensitivity to user-measured tag-size errors without requiring continuous motion or inter-pose trajectory estimation.

Experiments in simulation and on three real LiDAR--camera systems, covering mechanical and nonrepetitive LiDARs in clean and cluttered environments, support the effectiveness of these designs. Relative to single-pose calibration, multi-pose refinement reduced mean rotation and translation errors against the Kalibr-derived inter-camera reference on all three systems and improved rendered-map alignment. The tag-size sensitivity study further demonstrated the benefit of scale correction, while tests across five display types showed both the versatility of the approach and its limitations: small displays can provide insufficient reliable boundary measurements relative to LiDAR noise. These results establish ordinary displays as a practical calibration alternative when suitable screen geometry is available, with one static placement enabling rapid calibration and additional placements improving accuracy. Future work will investigate joint intrinsic and extrinsic refinement and improved handling of sparse, noisy display boundaries.


%

\ifCLASSOPTIONcaptionsoff
  \newpage
\fi



%

\makeatletter
\xpatchcmd{\IEEEbiographynophoto}
  {\vskip 4\baselineskip plus 1fil minus 0\baselineskip}
  {\vskip 4pt}
  {}{\PackageError{ScreenCalib}{Could not adjust IEEEbiographynophoto spacing}{Check the IEEEtran class version.}}
\makeatother
\raggedbottom

\begin{IEEEbiography}[{\includegraphics[width=1in,height=1.25in,clip,keepaspectratio]{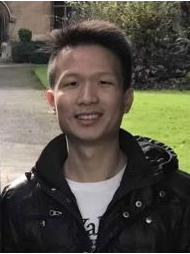}}]{Zuhao Zou}
received the B.Eng. degree in automatic control system engineering from the University of Sheffield in 2016 and the M.Sc. degree in computer science, specializing in computer vision and SLAM, from University College London in 2017. He received the Ph.D. degree from mechanical engineering at the University of Hong Kong and works in the MaRS Lab under Prof. Fu Zhang. His research spans SLAM, sensor fusion, loop detection, and place recognition. 
\end{IEEEbiography}

\begin{IEEEbiographynophoto}{Xiangcheng Hu}
received the B.Sc. degree from the North University of China in 2017 and the M.S. degree from Beihang University in 2020. He is pursuing the Ph.D. degree in electronic and computer engineering at The Hong Kong University of Science and Technology.
\end{IEEEbiographynophoto}

\begin{IEEEbiographynophoto}{Jianhao Jiao}
received the Ph.D. degree in robotics from The Hong Kong University of Science and Technology in 2021. He is a researcher at Lion Rock AI Lab, China Merchants Group. His research interests include autonomous navigation, multi-sensor SLAM, place recognition, and embodied intelligence.
\end{IEEEbiographynophoto}

\begin{IEEEbiography}[{\includegraphics[width=1in,height=1.25in,clip,keepaspectratio]{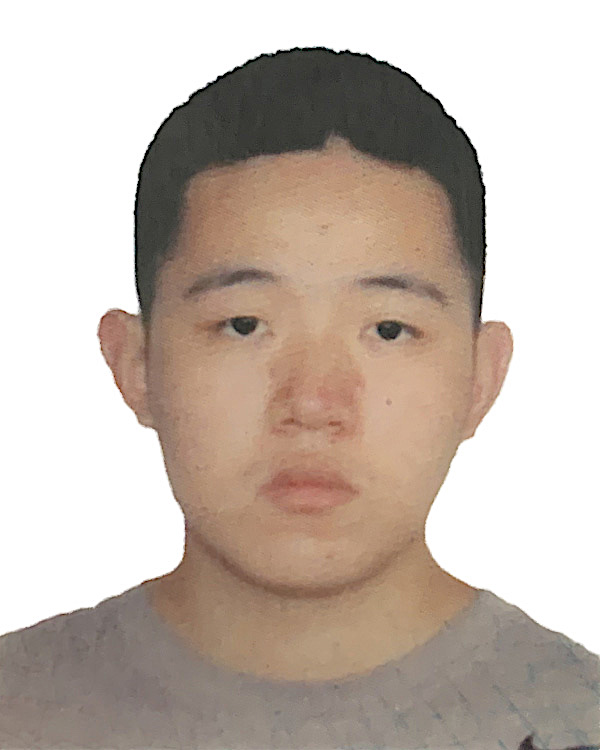}}]{Jin Wu}
received the Ph.D. degree in electronic and computer engineering from The Hong Kong University of Science and Technology in 2024 and the bachelor's degree in applied science from the University of Electronic Science and Technology of China. He is a Professor at the University of Science and Technology Beijing. His research interests include pose and attitude estimation, integrated navigation, sensor calibration, and multi-sensor fusion. He is an Technical Editor of \textsc{IEEE/ASME Transactions on Mechatronics} (TMECH), and an Associate Editor of \emph{International Journal of Robotics Research} (IJRR).
\end{IEEEbiography}

\end{document}